\pdfoutput=1

\documentclass[11pt]{article}

\usepackage[final]{coling}

\usepackage{times}
\usepackage{latexsym}

\usepackage[T1]{fontenc}

\usepackage[utf8]{inputenc}

\usepackage{microtype}

\usepackage{inconsolata}

\usepackage{graphicx}
\usepackage{algorithm}
\usepackage{algorithmic}
\usepackage{multirow}
\usepackage{svg}
\usepackage{amsmath}
\usepackage{amsfonts}
\usepackage{booktabs} 
\usepackage{colortbl} 
\usepackage[normalem]{ulem}
\useunder{\uline}{\ul}{}

\title{Large Language Model as a Teacher for Zero-shot Tagging at \\ Extreme Scales}

\author{
 \textbf{Jinbin Zhang\textsuperscript{1}},
 \textbf{Nasib Ullah\textsuperscript{1}},
 \textbf{Rohit Babbar\textsuperscript{1,2}},
\\
 \textsuperscript{1}Aalto University
 \textsuperscript{2}University of Bath
\\
 \small{
  {\{jinbin.zhang, nasibullah.nasibullah, 
 rohit.babbar\}@aalto.fi,}
  {rb2608@bath.ac.uk}
 }
}

\begin{document}
\maketitle
\begin{abstract}
Extreme Multi-label Text Classification (XMC) entails selecting the most relevant labels for an instance from a vast label set. Extreme Zero-shot XMC (EZ-XMC) extends this challenge by operating without annotated data, relying only on raw text instances and a predefined label set, making it particularly critical for addressing cold-start problems in large-scale recommendation and categorization systems. State-of-the-art methods, such as MACLR \cite{xiong-etal-2022-extreme}  and RTS \cite{zhang-etal-2022-structural-contrastive}, leverage lightweight bi-encoders but rely on suboptimal pseudo labels for training, such as document titles (MACLR) or document segments (RTS), which may not align well with the intended tagging or categorization tasks. On the other hand, LLM-based approaches, like ICXML \cite{zhu2024icxml}, achieve better label-instance alignment but are computationally expensive and impractical for real-world EZ-XMC applications due to their heavy inference costs. In this paper, we introduce LMTX\footnote{The Github link: https://github.com/xmc-aalto/LMTX} (\textbf{L}arge language \textbf{M}odel as \textbf{T}eacher for e\textbf{X}treme classification), a novel framework that bridges the gap between these two approaches. LMTX utilizes an LLM to identify high-quality pseudo labels during training, while employing a lightweight bi-encoder for efficient inference. This design eliminates the need for LLMs at inference time, offering the benefits of improved label alignment without sacrificing computational efficiency. Our approach achieves superior performance and efficiency over both LLM and non-LLM based approaches, establishing a new state-of-the-art in EZ-XMC.

\end{abstract}
\section{Introduction}
Extreme Multi-label Text Classification (XMC) is the task of assigning relevant labels to documents from an extensive label space, often comprising hundreds of thousands to millions of possible labels \cite{Bhatia16}. XMC is widely applied in real-world scenarios such as product-to-product recommendations, product search \cite{chang2021extreme}, labeling Wikipedia pages \cite{babbar2017dismec}, and categorizing Amazon products \cite{jiang2021lightxml}. Despite its widespread use, existing supervised XMC methods depend heavily on expert-annotated labels or user-annotated labels, with the label set fixed during both training and inference. Furthermore, supervised XMC faces two challenges. First, obtaining annotations is difficult due to the sheer scale of the label space, which makes it challenging for annotators to select relevant labels, often resulting in incomplete or missing labels \cite{qaraei2021convex, schultheis2021unbiased, schultheis2022missing, wydmuch2021propensity, jain2016extreme, schultheis2024generalized}. Second, the dynamic emergence of new labels, especially in cold-start scenarios adds further complexity. Conventional XMC methods are poorly equipped to handle unseen labels during inference, limiting their capacity to adapt to the evolving and dynamic nature of the label space.

There are two distinct settings for zero-shot extreme classification: (i) Generalized Zero-Shot Extreme Multi-label Learning (GZXML) \cite{gupta2021generalized}, which enables models to predict unseen labels but still relies on annotated training data, making it unsuitable for scenarios lacking labeled data, such as cold-start problems; and (ii) Extreme Zero-Shot Multi-label Text Classification (EZ-XMC) \cite{xiong-etal-2022-extreme, zhang-etal-2022-structural-contrastive}, which handles unseen labels without requiring any annotated data. In this work, we adopt the EZ-XMC setting to address cases where labeled data is unavailable, new labels emerge dynamically, and mainly focus on tagging application tasks.

Current EZ-XMC methods predominantly focus on training robust bi-encoders by leveraging pseudo-positive labels generated from the documents themselves. This approach enables the encoding of label texts into embeddings via a sentence encoder, facilitating efficient retrieval aligned with document embeddings. Crucially, this methodology eliminates the need for training datasets to encompass the entire label spectrum. For instance, MACLR \cite{xiong-etal-2022-extreme} constructs instance-pseudo label pairs using (content, title) combinations from documents, while RTS \cite{zhang-etal-2022-structural-contrastive} randomly splits documents and selects two spans to form such pairs (Figure \ref{fig:blt_diff}). However, these methods often overlook the direct semantic alignment between the document and pseudo-label pairs. For instance, a segment generated by the RTS might not be relevant to another segment if they are located too far apart within the same document.  Moreover, the pseudo-labels may not adequately reflect the domain of the predefined label set, leading to a mismatch between the target task and the generated training pairs.

Large Language Models (LLMs) have recently exhibited remarkable reasoning and zero-shot capabilities across diverse NLP tasks \cite{bonifacio2022inpars, saad2023udapdr, ma2023zero, qin2023large, sun2023chatgpt, dai2023uncovering, hou2023large, sachan2023questions, sachan-etal-2022-improving}.  Nevertheless, only a few notable exceptions \cite{zhu2024icxml, xu2023dense, liu2024xmc} have been explored in the context of XMC problems. This limited adoption is primarily due to the substantial computational overhead associated with deploying LLMs, especially given the large search space typical of XMC tasks. Additionally, the inference phase for XMC problems can become prohibitively expensive when using heavy LLM models. To address this limitation, we propose a novel relevance assessment strategy that leverages an LLM to judiciously select high-quality pseudo labels from a curated label set for each document. This approach enables the training of a lightweight bi-encoder model that inherits the LLM's knowledge while avoiding the inference-time computational burden.
Our contributions can be summarized as follows:
\begin{figure}[]
  \centering
  \includegraphics[width=0.95\linewidth]{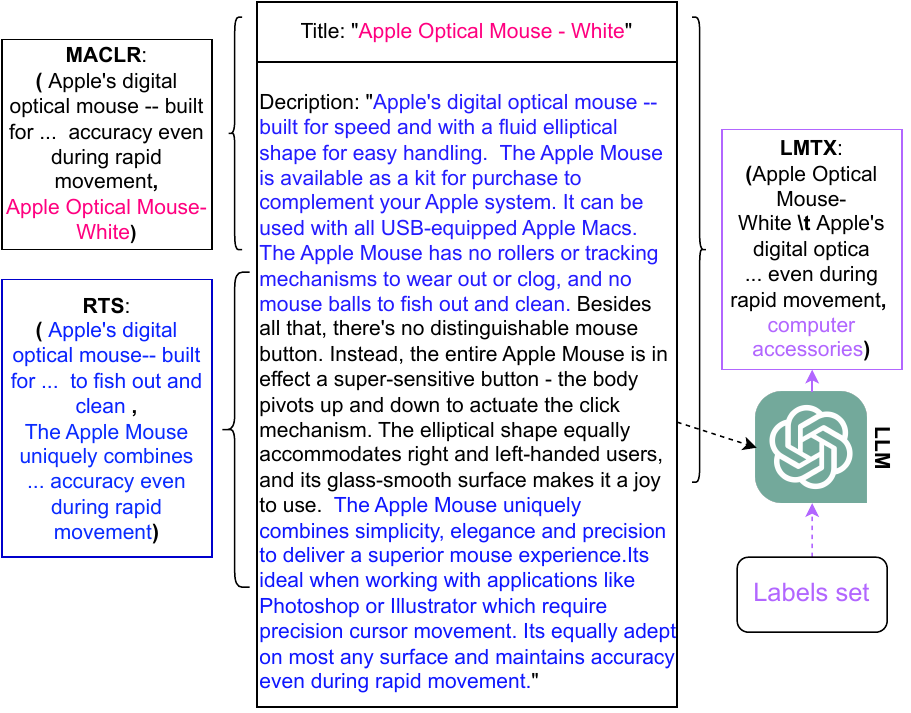}
  \caption{An example of how to construct training pairs using state-of-the-art methods MACLR \cite{xiong-etal-2022-extreme} and RTS \cite{zhang-etal-2022-structural-contrastive}. MACLR utilizes the `Title' of a document to generate pseudo labels, while the `Description' serves as the training document. Conversely, RTS forms its training pairs by selecting two random segments from the `Description'. Differently, our proposed model, LMTX, adopts a more refined approach. It selects `computer accessories' as a pseudo positive label from a predefined set, a choice validated by the LLM model.}
  \label{fig:blt_diff}
\end{figure}

\begin{itemize}

\item LMTX introduces a novel training approach for bi-encoders, emphasizing a curriculum-based method that dynamically adjusts based on the relevance feedback from an LLM by leveraging its zero-shot learning abilities.
\item The proposed LMTX requires less training data because there is a higher correlation between the pseudo-labels and documents, resulting in higher-quality training pairs. Consequently, our approach achieves better performance while maintaining similar or reduced training time compared to traditional methods for some large datasets.
\item The proposed LMTX enables the lightweight deployment by using only the bi-encoder to generate embeddings for documents and labels during the prediction. LLM models are not involved in the prediction process. LMTX significantly outperforms current state-of-the-art methods for the tagging task, demonstrating comprehensive advancements in performance metrics. 
\end{itemize}

\begin{figure*}[]
  \centering
  \includegraphics[width=0.88\linewidth]{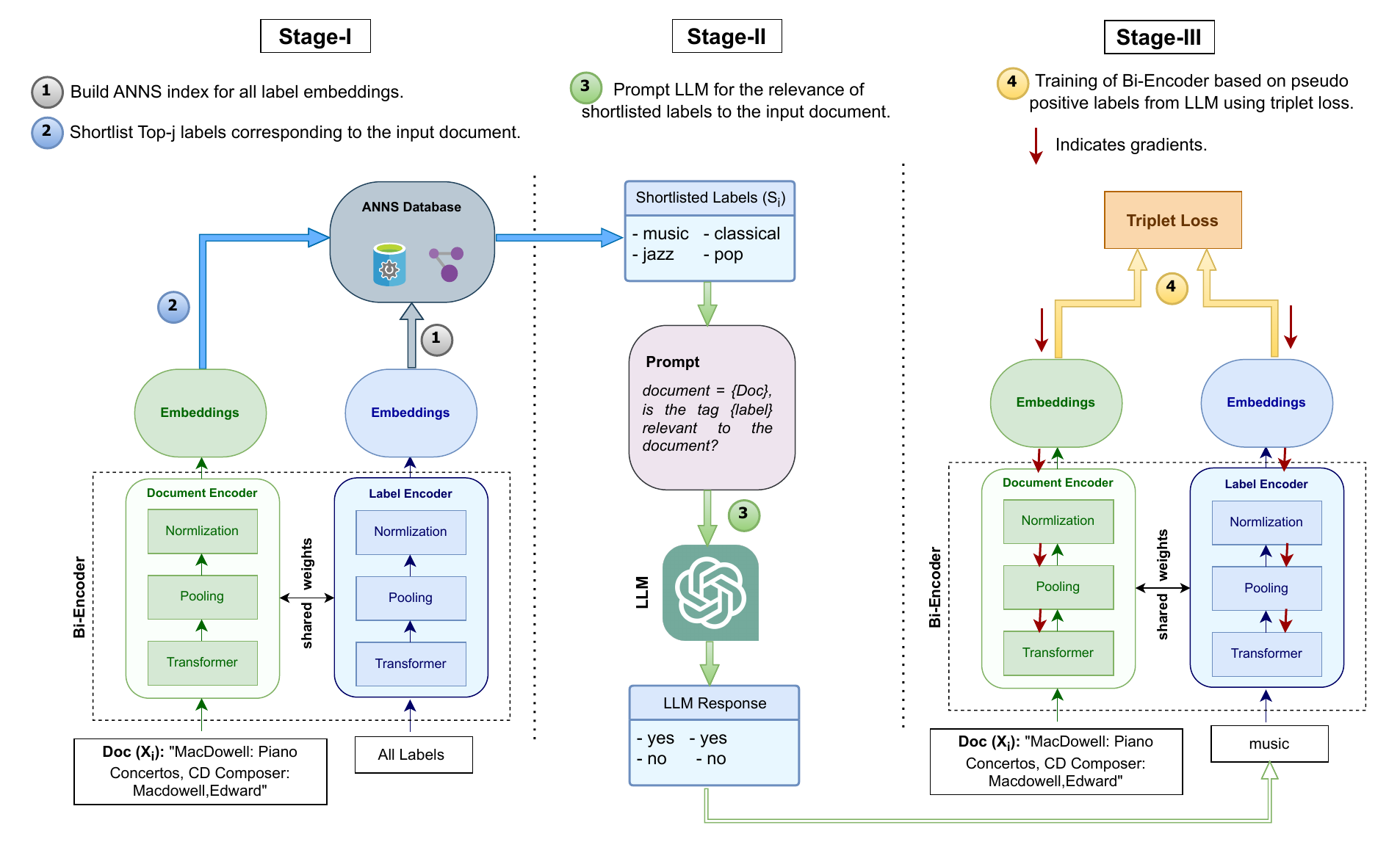}
  \caption{The process of getting feedback from LLM model for training the bi-encoder. First, for a given document, the (pre) trained bi-encoder and ANNS are employed to create a short-list of potential labels. Next, the LLM assesses the relevance between the labels in this shortlist and the document. Finally, the selected labels are utilized to further train the bi-encoder.}
  \label{fig:framework}
\end{figure*}

\section{Background}
\textbf{Problem Definition:} Let's denote $X_i \in \mathcal{X}$ as the text for an instance in a particular domain; i.e., $X_i$ could be the textual description for a product on Amazon.
Unlike the supervised XMC, the key characteristic of the EZ-XMC setting is that we do not have the corresponding well-annotated labels $Y_i$ for each training instance $X_i$. 
However, besides having the original text of instances $\{X_i\}_{i=1}^{N}$, we also have access to  the predetermined labels along with their texts, i.e., we have $\{l_k\}_{k=1}^{L}$. %where $l_{k} \in \mathcal{X}$.
We refer to this collection of predetermined labels as the ``labels set''.
The goal of EZ-XMC, which is the one that we consider in this paper, is to assign the document $X_i \in \mathcal{X}$ to set of labels $ \{l_{j}\} \subseteq \{l_k\}_{k=1}^{L}$ that are relevant to the document. To achieve this objective, the task requires learning a mapping function from text to embeddings for both $\{X_i\}_{i=1}^{N}$ and $\{l_k\}_{k=1}^{L}$, so that the $\{l_k\}_{k=1}^{L}$ can be retrieved in the same space as $\{X_i\}_{i=1}^{N}$  by comparing their embedding similarity. The mapping function is denoted as $\mathcal{E}_{\theta }: \mathcal{X} \rightarrow \mathbb{S}^{D-1} $, where $\theta$ represents the training parameters, $\mathcal{E}$ represents the encoder for documents and labels, and $\mathbb{S}^{D-1}$ is the $D$-dimensional unit sphere. The mapping function is typically implemented as a bi-encoder, where both the text of documents and labels are embedded within $\mathbb{S}^{D-1}$.

\noindent \textbf{Bi-Encoder Model:} We employ a bi-encoder architecture, $\mathcal{E}_{\theta}$, to generate embeddings for both document and label text. The model consists of two encoders with shared weights: one for documents and another for labels. The document and label embeddings are represented as $\mathcal{E}_{\theta}(X_i)$ and $\mathcal{E}_{\theta}(l_k)$, respectively, where $X_i$ is the document and $l_k$ is the label text. The relevance score between document $X_i$ and label $l_k$ is computed via cosine similarity between their embeddings. The bi-encoder we use is based on the Distill-BERT transformer \cite{sanh2019distilbert} and depicted in Figure \ref{fig:framework}.

\section{Training the Bi-encoder from the Feedback of LLM}
\noindent 
\textbf{Training Process Overview:} Our  
methodology adopts an iterative framework, encompassing three distinct stages within each cycle. 
Initially, we embed all documents and labels, subsequently constructing an Approximate Nearest Neighbor Search (ANNS \cite{malkov2018efficient}) over the label embeddings to retrieve a refined set of label candidates for each document. In the second stage, the LLM is deployed to scrutinize these candidates, effectively identifying pseudo positive labels.
The final stage involves training the bi-encoder model using the labels identified in the preceding stage. Figure \ref{fig:framework} illustrates the mechanism through which the bi-encoder incorporates feedback from the LLM and progresses through training regimen.

\noindent \textbf{Data Embedding \& Shortlist Generation (stage-I)}: The LLM model demonstrates zero-shot ability in determining relevance between two text segments \cite{ma2023zero}. However, this approach encounters %practical 
 challenges when applied to a vast array of labels, as in our context. Specifically, the computational complexity involved in assessing the relevance between each document and every label in a large set becomes formidable, being $\mathcal{O}(NL)$ in complexity. This can be quite prohibitive, even  for a dataset with a moderate number  $(\mathcal{O}(10^3))$ of instances and labels. To mitigate this, our strategy involves condensing the label space presented to the LLM. We utilize (pre) trained bi-encoder to process the document and label text into embeddings and utilize ANNS to efficiently select the top-j most relevant labels for each document. These selected labels, denoted as $S_i = \{l_{i1}, l_{i2}, ..., l_{ij} \}$, constitute a focused subset for subsequent processing. 

\noindent \textbf{LLM Model as a Teacher (stage-II)}: Once we obtain the label shortlist $S_i$ for the $i$-th document, we can employ the LLM as a teacher to determine the relevance between the document and the top-$j$ labels in a shortlist. 
Let $X_i$ denote a particular document and $l_{ik}$ be its $k$-th label in the shortlist. %``labels set''
To assess the relevance between $X_i$ and $l_{ik}$, we instruct the LLM with the question, ``document = \{$X_i$\}, is the tag \{$l_{ik}$\} relevant to the document? answer yes or no''. If the LLM outputs ``Yes'', we consider $l_{ik}$ to be relevant to $X_i$. Conversely, if the model outputs ``No'', we consider $l_{ik}$ as an unrelated label and discard it. We keep all the labels from the shortlist that received a positive feedback (``yes'') from the LLM. Then, we use these selected relevant labels to train the bi-encoder model. 
A detailed discussion of different prompts used for the LLM can also be found in Appendix \ref{prompts_llm}.

\begin{algorithm}[]
\caption{Training the bi-encoder with the feedback from LLM teacher (LMTX)}
\small
\label{alg:algorithm}
\begin{flushleft}
\textbf{Input}: Initial bi-encoder $\mathcal{E}_\theta$, LLM model $\mathcal{M}_{LLM}$, data instances $\{X_i\}_{i=1}^{N}$, labels set $\{l_k\}_{i=1}^{L}$, dev set instances $\{X_j\}$, and number of cycles $T$ \\
\textbf{Output}: Trained bi-encoder $\mathcal{E}_\theta$ 

\end{flushleft}
\begin{algorithmic}[1]
\STATE  $c=0$.
\WHILE{$c < $ $T$}
    \STATE Compute $\mathcal{E}_\theta(X_{i})$, $\mathcal{E}_\theta(l_{k})$  for all $\{X_i\}_{i=1}^{N}$ and $\{l_k\}_{i=1}^{L}$
    \STATE  Retrieve top labels $S_{i}=ANNS(\mathcal{E}_\theta(X_{i}),\mathcal{E}_\theta(l_{k})_{k=1}^{L})$ for each $X_{i}$        
    \STATE Fetch pseudo positive labels $P_{i}^{+} = \mathcal{M}_{LLM}(X_{i},S_{i}) $ for all $X_{i}$

    \FOR{i=0 to N\_batches} 
    \STATE \raggedright Sample a mini\_batch  $B_{i}=\{X_{i},P_{i}^{+} \}$  where, $\lvert B_{i} \rvert = m$ %// $N = N\_batches * |B_{i}| $ 

    \STATE Update $\mathcal{E}_\theta$ using mini-batch $B_i$, loss $\mathcal{L}$ and AdamW optimizer.
    \ENDFOR
    \STATE Evaluate $\mathcal{E}_\theta$  with $\mathcal{M}_{LLM}$ on the dev set $\{X_j\}$ and obtain $P@1$ over pseudo labels.
    \IF {$P@1$ does not improve on dev dataset}
    \STATE Stop training cycle
    \ENDIF
    \STATE $c = c + 1$

\ENDWHILE
\STATE \textbf{return} model $\mathcal{E}_\theta$
\end{algorithmic}
\end{algorithm}

\noindent \textbf{Training Bi-Encoder with Pseudo Positive Labels (stage-III)}: To train the bi-encoder, we follow the training procedure in \cite{dahiya2023ngame}. 
Out of the labels identified by the LLM as the pseudo positives, we choose only one of the pseudo positive labels for each document during the training process. 
This is shown to help in achieving faster convergence in the earlier work \cite{dahiya2023ngame}. 
Regarding the negatives, which we need to compute the instance-wise loss, we use in-batch negative sampling, in which the negatives for a document come from the pseudo positive labels of other documents in the same batch. Our analysis in Section \ref{sec:ablation} shows that using labels which are rejected by the LLM, as hard negatives, leads to degradation in prediction performance.

For the label $l_k$, the predicted relevance score between document $X_i$ and $l_k$ is computed through the cosine similarity $\langle \mathcal{E}_\theta(X_i), \mathcal{E}_\theta(l_k) \rangle$, and 
triplet loss is used to train the bi-encoder \cite{Schroff_2015_CVPR,Manmatha2017SamplingMI,dahiya2023ngame}:
\begin{equation}
\small
 \mathcal {L} = \sum_{i=1 }^{N}\sum_{k'}[\langle \mathcal{E}_\theta(X_i), \mathcal{E}_\theta(l_{k'})\rangle -  \langle \mathcal{E}_\theta(X_i) , \mathcal{E}_\theta(l_p)\rangle + \gamma]_+\label{loss}
\end{equation}
where $\gamma$ is the margin, the $k'$ stands for the index of hard negative labels from the mini batch,  $l_{k'}$ and $l_p$ correspond to the text of the negative labels and the pseudo positive label.

As training progresses, the bi-encoder gradually improves, leading to an enhancement in the quality of labels within the shortlist and increased relevance to the corresponding document. 
During training, we evaluate the model on the development dataset and choose the best one based on performance evaluated by the LLM since under the EZ-XMC setting one does not have  access to annotated ground-truth labels.
If there is no performance improvement on the development set, training is halted, so the number of cycles is actually determined by the performance on the development dataset. The pseudo code of the proposed algorithm LMTX, for training the bi-encoder model with feedback from LLM, is presented in Algorithm \ref{alg:algorithm}.

\noindent \textbf{Inference:} The model's inference procedure is analogous to the formation of the shortlist during training, as depicted in Stage-I of Figure \ref{fig:framework}. We build the MIPS \cite{johnson2019billion} over these label embeddings, which implements the efficient maximum inner product search. For each document, we employ its embedding as a query to retrieve the top-$m$ labels, which ultimately
serve as the predicted results.
The use of MIPS\footnote{https://github.com/facebookresearch/faiss} in the inference process ensures a sublinear time complexity for each instance. The label embedding extraction and construction of MIPS index are performed just once, hence amortizing the cost of this step.

\begin{table}[ht]\small
\centering
\resizebox{\linewidth}{!}{%
\begin{tabular}{l|cccc}
\toprule
\textbf{Dataset}  & $\boldsymbol{N}$ & $\boldsymbol{N_{test}}$ & $\boldsymbol{N_{label}}$ & $\boldsymbol{L_{N}}$ \\ 
\midrule
EURLex-4K         &  15,511 &  3,803 & 3,956 & 20.79 \\
Wiki10-31K        &  14,146 &  6,616 & 30,938 & 8.52 \\
AmazonCat-13K     &  1,186,239 & 306,782  &  13,330 & 448.57 \\
LF-WikiSeeAlso-320K   & 693,082  & 177,515  & 312,330 & 4.67 \\
LF-Wikipedia-500K & 1,813,391  &  783,743 &  501,070 & 24.75 \\ 
\bottomrule
\end{tabular}}
\caption{Statistical overview of the datasets. $N$: total number of training samples, $N_{test}$: number of test samples, $N_{label}$: total number of unique labels, $L_{N}$: average number of samples per label.}
\label{statistics}
\end{table}

\section{Experiments}

\textbf{Datasets and Evaluation Metrics:} We utilized five tagging datasets for evaluation: EURLex-4k, Wiki10-31k, and AmazonCat-13K were obtained from the XLNet-APLC repository\footnote{https://github.com/huiyegit/APLC\_XLNet}, while the remaining datasets were downloaded from the extreme classification repository\footnote{http://manikvarma.org/downloads/XC/XMLRepository.html}.
Table \ref{statistics} provides comprehensive statistical information for all datasets.  To optimize computational resources, we constrained the training data for AmazonCat-13K, LF-WikiSeeAlso-320K, and LF-Wikipedia-500K to 30,000 documents each. In contrast, baseline models utilize the entire dataset.

\begin{table*}[ht]
\centering
\resizebox{\textwidth}{!}{%
\begin{tabular}{lccccccc|ccccccc}
\toprule
Method & P@1  & P@3  & P@5  & R@1  & R@3  & R@5 & R@10  & P@1  & P@3  & P@5  & R@1  & R@3  & R@5 & R@10 \\ \midrule
\rowcolor{gray!30} 
& \multicolumn{7}{c}{\textbf{EURLex-4K}} & \multicolumn{7}{c}{\textbf{Wiki10-31K}} \\ \midrule
Glove                   & 1.66  & 1.11  & 1.04  & 0.37  & 0.73  & 1.08  & 1.88  & 3.87  & 3.11  & 2.87  & 0.24  & 0.57  & 0.89  & 1.48\\
SentBERT & 8.52 & 7.70  & 6.83  & 1.70  & 4.54  & 6.69 & 10.20 & 9.39 & 6.93  & 5.81  & 0.60  & 1.31 & 1.81 & 2.70 \\
SimCSE      & 5.86  & 4.44  & 3.85  & 1.20  & 2.86  & 3.93  & 6.12  & 23.55  & 17.21  & 14.01  & 1.42  & 3.07  & 4.13 & 6.01  \\
MPNet & 10.81 & 8.65 & 7.21 & 2.27 & 5.28 & 7.28 & 10.85 & 44.82 & 29.18 & 22.38 & 2.63 & 5.12 & 6.52 & 8.89 \\
Msmacro-distilbert     & 15.91 & 9.89 & 7.81  & 3.33  & 6.16 & 8.08 & 11.22 & 54.17$^{\S}$ & 33.44$^{\S}$  & 25.38$^{\S}$  & 3.18$^{\S}$  & 5.82$^{\S}$ & 7.32$^{\S}$ & 9.70$^{\S}$ \\
RTS       & 30.58$^{\S}$ & 21.54$^{\S}$  & 17.73$^{\S}$  & 6.19$^{\S}$  & 13.01$^{\S}$  & 17.72$^{\S}$  & 25.34$^{\S}$  & 47.73  & 31.03  & 23.65  & 2.81  & 5.41  & 6.84  & 9.12 \\ \midrule
LMTX       & \textbf{47.28$^\dag$ } & \textbf{29.34$^\dag$ }  & \textbf{21.98$^\dag$ }  & \textbf{9.6$^\dag$ }  & \textbf{17.68$^\dag$ }  & \textbf{21.96$^\dag$ }  & \textbf{28.44$^\dag$ }  & \textbf{57.89$^\dag$ } & \textbf{38.00$^\dag$ } & \textbf{29.09$^\dag$ } & \textbf{3.41$^\dag$ } & \textbf{6.68$^\dag$ } & \textbf{8.46$^\dag$ } & \textbf{11.14$^\dag$ }  \\
 \midrule
 \rowcolor{gray!30} 
& \multicolumn{7}{c}{\textbf{AmazonCat-13K}} & \multicolumn{7}{c}{\textbf{LF-WikiSeeAlso-320K}} \\ \midrule
Glove                   & 4.83  & 3.89  & 3.42  & 0.99  & 2.46  & 3.67  & 6.05  & 3.86  & 2.76  & 2.21  & 2.12  & 4.11 & 5.22 & 6.95 \\
SentBERT & 5.21 & 4.22  & 3.68  & 0.99  & 2.34  & 3.37 & 5.35 & 1.71 & 1.27  & 1.06  & 1.08  & 2.16 & 2.90 & 4.17  \\
SimCSE      & 2.84  & 2.60  & 2.42  & 0.52  & 1.41  & 2.17  & 3.75  & 9.03  & 6.64  & 5.22  & 4.99  & 9.89  & 12.34 & 15.93  \\
ICT      & 15.52  & 10.48  & 8.34  & 2.91  & 5.93  & 7.86  & 11.04  & 10.76  & 10.05  & 8.12  & 6.12  & 14.32  & 18.05 & 23.01  \\
MPNet & 18.01 & 12.84 & 10.51 & 3.63 & 7.68 & 10.48 & 15.73 & 13.75 & 11.93 & 9.58 & 8.14 & 17.77 & 22.21 & 28.11  \\
MACLR & 10.66 & 6.75 & 5.14 & 1.98 & 3.79 & 4.81 & 6.35 & 16.31 & 13.53 & 10.78 & 9.71 & 20.39 & 25.37 & 32.05  \\
Msmacro-distilbert     & 16.36 & 10.96 & 8.68  & 3.29  & 6.62 & 8.73 & 12.23 & 14.93 & 12.65  & 10.08  & 8.99  & 19.25 & 23.99 & 30.19 \\
RTS       & 18.89$^{\S}$ & 13.59$^{\S}$  & 11.07$^{\S}$  & 3.69$^{\S}$  & 8.03$^{\S}$  & 10.97$^{\S}$  & 16.20$^{\S}$  & 18.64  & \textbf{15.14}  & \textbf{12.07}  & 10.86  & \textbf{22.68}  & \textbf{28.29} & \textbf{35.47} \\ \midrule
LMTX       & \textbf{25.91$^\dag$ } & \textbf{17.08$^\dag$ }  & \textbf{13.12$^\dag$ }  & \textbf{5.53$^\dag$ }  & \textbf{10.77$^\dag$ }  & \textbf{13.60$^\dag$ }  & \textbf{17.84$^\dag$ }  & \textbf{19.11} & 14.00 & 10.95 & \textbf{11.41} & 21.38 & 26.10 & 32.44 \\
\midrule
\rowcolor{gray!30} 
& \multicolumn{7}{c}{\textbf{LF-Wikipedia-500K}} & \multicolumn{7}{c}{\textbf{}} \\ \midrule
Glove                   & 2.19  & 1.52  & 1.23  & 0.85  & 1.66  & 2.18  & 3.10 &   &   &   &   &   & \\
SentBERT & 0.17 & 0.15  & 0.13  & 0.05  & 0.13  & 0.18 & 0.30  &  &  &   &   &   &  \\
SimCSE      & 14.32  & 6.84  & 4.55  & 4.24  & 8.03  & 11.26  & 14.35 &  &   &   &   &   &   \\
ICT      & 17.74  & 9.67  & 7.06  & 7.35  & 11.60  & 13.84  & 17.19 &  &   &   &   &   &   \\
MPNet & 22.46 & 12.87 & 9.49 & 8.74 & 14.07 & 16.76 & 20.64 &  &  &  &  &  &  \\
Msmacro-distilbert     & 21.62 & 12.75 & 9.52  & 8.27  & 13.81 & 16.68 & 20.89 &  &  &   &   &   &  \\
MACLR      & 28.44  & 17.75  & 13.53  & 10.40  & 18.16  & 22.38  & 28.52 &  &   &   &   &   &   \\
RTS       & 30.67 & 19.03  & 14.34  & 10.58  & 18.48  & 22.51  & 28.23  &  &   &   &   &   &  \\ 
\midrule
LMTX       & \textbf{40.25} & \textbf{23.00}  & \textbf{16.81}  & \textbf{13.65}  & \textbf{22.15}  & \textbf{26.16}  & \textbf{31.61}  &  &   &   &   &   &  \\
\bottomrule
\end{tabular}%
}
\caption{Comparison of LMTX model with state-of-the-art EZ-XMC methods. The symbol $\dag$ indicates a statistically significant improvement over the best baseline model (paired t-test with $p\leq 0.01$) and the symbol $\S$ represents the best baseline model.}
\label{tab:LMTX_performance}
\end{table*}

We employ the commonly used evaluation metrics \cite{pmlr-v89-reddi19a,ChangJYTZZKHSIS21, zhang-etal-2022-structural-contrastive} for the EZ-XMC setting: $Precision@k$ and $Recall@m$. Further details on the evaluation metrics and implementation can be found in the Appendix 
\ref{app:evaluation_metrics} and \ref{app:implementation} respectively.

\noindent \textbf{Baselines:} We have incorporated state-of-the-art EZ-XMC models as our baselines. The baseline contains unsupervised pseudo-labels methods: MACLR \cite{xiong-etal-2022-extreme} and RTS \cite{zhang-etal-2022-structural-contrastive}. Unsupervised pre-trained embeddings and encoders: GloVe \cite{pennington-etal-2014-glove}, Inverse Cloze Task (ICT) \cite{lee2019latent} and MPNet \cite{song2020mpnet}. Sentence matching: SentBERT \cite{reimers2019sentence} and SimCSE \cite{gao2021simcse}. Pre-trained retrieval bi-encoder: Msmarco-distilbert \cite{reimers2021curse}. LLM-based methods: ICXML \cite{zhu2024icxml}. To assess the baseline performance of LF-WikiSeeAlso-320K and LF-Wikipedia-500k, we obtained the results from \cite{zhang-etal-2022-structural-contrastive}. As for the other baselines, we acquired their performance by executing the respective baseline. 

\noindent \textbf{Comparison with standard baselines:} In Table \ref{tab:LMTX_performance}, we present a comparative analysis of our model's performance against other models. Notably, our LMTX model demonstrates substantial improvements in both $Precision@m$ \& $Recall@m$, especially for datasets like EURLex-4k, Wiki10-31k, AmazonCat-13k, and LF-Wikipedia-500k. Particularly striking are the results in LF-Wikipedia-500k and AmazonCat-13K, where our model shows an increase of 31\% and 37\%, respectively, for $P@1$. In addition, our results on LF-WikiseeAlso-320k are competitive with those of the leading models, despite the unique nature of this task, which focuses on identifying related Wikipedia titles rather than traditional tagging. 
Moreover, Table \ref{tab:training time} presents a comparison of the training time and computational resources required for LMTX relative to other methods, further underscoring the efficiency of our approach. These results strongly indicate that our approach is both computationally efficient and highly effective in zero-shot scenarios, capable of addressing diverse tagging and categorization tasks with state-of-the-art performance.

\begin{table*}[]
\small
\centering
\resizebox{\linewidth}{!}{%
\begin{tabular}{c|c|cccc|cc}
\toprule
\multicolumn{1}{c|}{\textbf{Dataset}} & \textbf{Methods} & \multicolumn{1}{c}{\textbf{P@1}} & \multicolumn{1}{c}{\textbf{P@5}} & \multicolumn{1}{c}{\textbf{R@1}} & \multicolumn{1}{c|}{\textbf{R@5}} & \multicolumn{1}{c}{\textbf{Inf. time}} & \textbf{GPUs} \\ 
\midrule
\multirow{4}{*}{EURLex-4K} & ICXML-WizardLM-13B & 2.21 & 2.28 & 0.5 & 2.39 & 16.46 & 1x(A100-40GB) \\
 & ICXML-Vicuna-33B & 7.47 & 6.05 & 1.64 & 6.15 & 35.28 & 2x(A100-40GB) \\
 & ICXML-Llama3-70B & 19.14 & 16.51 &  3.85 & 16.27 & 21.32 & 4x(A100-80GB) \\
 & LMTX-DistilBERT-66M & \textbf{47.28} & \textbf{21.98} & \textbf{9.6} & \textbf{21.96} & 0.019 & 1x(A100-40GB) \\ 
 \midrule
\multirow{4}{*}{\begin{tabular}[c]{@{}l@{}}LF-WikiSeeAlso-320K \end{tabular}} & ICXML-WizardLM-13B & 4.38 & 2.94 & 2.74 & 8.28 & 19.48 & 1x(A100-40GB) \\
 & ICXML-Vicuna-33B & 6.6 & 4.12 & 3.76 & 11.29 & 24.29 & 2x(A100-40GB) \\
 & ICXML-Llama3-70B & \textbf{26.13} & \textbf{13.93} & \textbf{13.54} & \textbf{31.02} & 15.53 & 4x(A100-80GB) \\
 & LMTX-DistilBERT-66M & 19.22 & 10.94 & 11.15 &  26.01 & 0.032 & 1x(A100-40GB) \\ 
 \bottomrule
\end{tabular}%
}
\caption{
Performance comparison of LMTX and ICXML on EURLex-4K and LF-WikiSeeAlso-320K datasets. The table shows precision and recall metrics, inference time (in hours), and the number of GPUs used. Results for LF-WikiSeeAlso-320K are averaged over two 3500-sample subsets.}
\label{tab:llm-methods}
\end{table*}

\noindent \textbf{Comparison with LLM-based baseline:} We compared our results  against ICXML \cite{zhu2024icxml} (only LLM baseline for EZ-XMC) using various LLM models as shown in Table \ref{tab:llm-methods}. On EURLex-4K, LMTX significantly outperforms all ICXML variants, achieving a P@1 of 47.28 versus 19.14. On LF-WikiSeeAlso-320K, LMTX demonstrates superior performance compared to models up to 33B in size, with an insignificant performance drop relative to the substantial difference in model size (70B vs 66M).
Crucially, LMTX achieves these results with significantly reduced computational demands and substantially lower inference times, enabling more scalable real-world deployment.

\section{Ablations and Comprehensive Analysis} 
\label{sec:ablation}
\noindent \textbf{Evaluating Teacher Models (Analyzing Open-Source LLMs):} We evaluate open-source LLMs as potential teacher models. Table \ref{tab:teacher} presents the performance results using different recently released LLM model families, with same parameters (13B). 

Our analysis reveals that WizardLM outperforms other models on the AmazonCat-13K and LF-WikiSeeAlso-320K datasets, while Llama2 demonstrates improved performance over WizardLM on the LF-Wikipedia-500K dataset. These findings underscore the versatility of our proposed method, which is not confined to a single LLM model. This flexibility enables selection of the most appropriate teacher model to achieve optimal performance.

\begin{table}[h]
\small
\centering
\resizebox{\linewidth}{!}{%
\begin{tabular}{c|c|cccc}
\toprule
\textbf{Dataset} & \textbf{LLM Model} & \textbf{P@1} & \textbf{P@5} & \textbf{R@1} & \textbf{R@5} \\
\midrule
\multirow{3}{*}{\parbox{2cm}{\centering AmazonCat-13K}}
 & WizardLM & \textbf{25.91} & \textbf{13.12} & \textbf{5.53} & \textbf{13.60} \\
 & Vicuna   & 25.01 & 12.70 & 5.24 & 12.95 \\
 & Llama2   & 25.21 & 12.76 & 5.34 & 13.22 \\
\midrule
\multirow{3}{*}{\parbox{2cm}{\centering LF\\WikiSeeAlso-\\320K}}
 & WizardLM & \textbf{19.11} & 10.95 & \textbf{11.41} & 26.10 \\
 & Vicuna   & 17.76 & \textbf{11.07} & 10.91 & \textbf{26.58} \\
 & Llama2   & 17.59 & 10.64 & 10.46 & 25.27 \\
\midrule
\multirow{3}{*}{\parbox{2cm}{\centering LF\\Wikipedia-\\500K}}
 & WizardLM & 40.25 & 16.81 & 13.65 & 26.16 \\
 & Vicuna   & 39.37 & 16.78 & 13.47 & 26.04 \\
 & Llama2   & \textbf{41.67} & \textbf{17.20} & \textbf{14.37} & \textbf{26.86} \\
\bottomrule
\end{tabular}
}
\caption{Comparison of different LLM models as teacher.}
\label{tab:teacher}
\end{table}

\begin{figure}[ht]
  \centering
  \includegraphics[width=\linewidth]{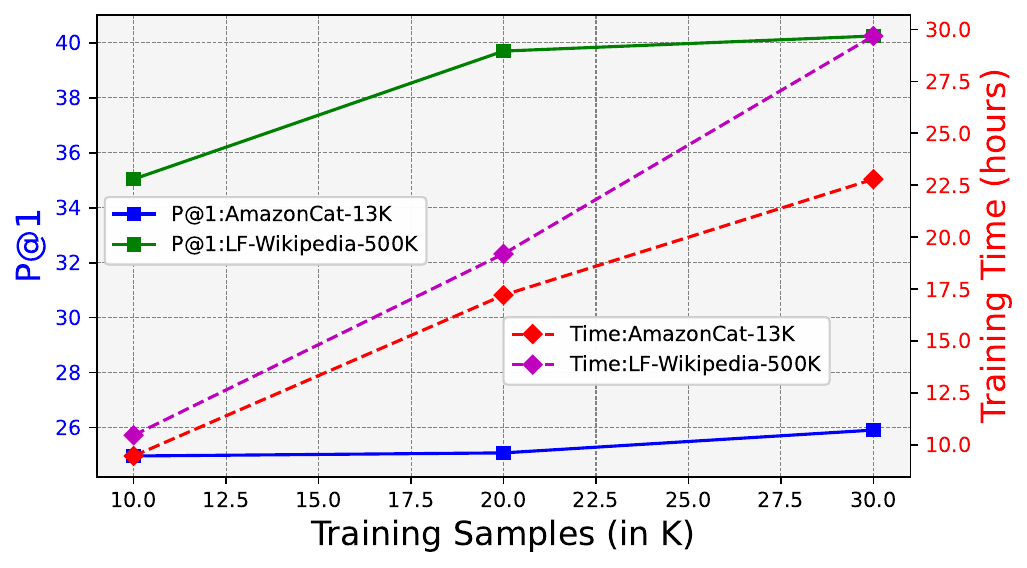}
  \caption{ Effect of training sample size on LMTX performance and training time.}
  \label{fig:ablation1}
\end{figure}

\noindent \textbf{Optimizing Training Efficiency: Impact of Sample Size on Performance and Training Time} To enhance the efficiency and cost-effectiveness of LMTX, particularly when incorporating LLM-based teacher models in large datasets, we trained the bi-encoder using a subset of the training dataset. As depicted in Figure \ref{fig:ablation1}, we systematically investigated the impact of reducing the number of documents on both final performance and training time by randomly sampling data from the entire dataset.
The results show that increasing the number of training samples improves model performance, as evidenced by higher P@1 scores. However, this improvement is accompanied by a significant increase in training time, underscoring the necessity of balancing performance gains with the corresponding training time.
\begin{table}[h]
\small
\centering
\resizebox{\linewidth}{!}{%
\begin{tabular}{p{1.8cm}|c|cccc}
\toprule
\centering \textbf{ Dataset} & \textbf{Initialization} & \textbf{P@1} & \textbf{P@5} & \textbf{R@1} & \textbf{R@5} \\ 
\midrule
\multirow{2}{*}{\parbox{2cm}{\centering AmazonCat-13K}} 
    & RTS-SI & 17.87 & 10.35 & 3.57  & 10.61 \\
    & LMTX   & \textbf{25.91} & \textbf{13.12} & \textbf{5.53}  & \textbf{13.60} \\ 
\midrule
\multirow{2}{*}{\parbox{2.5cm}{LF-
WikiSee\\Also-320K}} 
    & RTS-SI & 14.82 & 8.89  & 8.41  & 21.02 \\
    & LMTX   & \textbf{19.11} & \textbf{10.95} & \textbf{11.41} & \textbf{26.10} \\ 
\bottomrule
\end{tabular}
}
\caption{Performance comparison across datasets with consistent initialization. RTS-SI uses same initialization as ours.}
\label{tab:initialization}
\end{table}

\noindent \textbf{Assessing Initialization Robustness:} The choice of initialization influences both the quality of the initial label shortlist and the subsequent training process of the bi-encoder. To isolate the effects of our method from potential biases due to initialization advantages, we applied identical initialization procedures to both our approach and the best non-LLM baseline RTS \cite{zhang-etal-2022-structural-contrastive}. The results, as presented in Table \ref{tab:initialization}, demonstrate that our method consistently outperforms the baseline, even when identical initialization is applied. 
We also include the performance of the initialized bi-encoder model, msmarco-distilbert-base-v4\footnote{https://huggingface.co/sentence-transformers/msmarco-distilbert-base-v4}, in Table \ref{tab:LMTX_performance}. The results demonstrate that training with the proposed method improves the bi-encoder model, making it outperform the initialized model on XMC problems. These results indicate that our method's efficacy stems from intrinsic improvements in the learning process rather than initialization advantages, underscoring its robustness and broad applicability.

\begin{figure}[ht]
  \centering
  \includegraphics[width=0.8\linewidth]{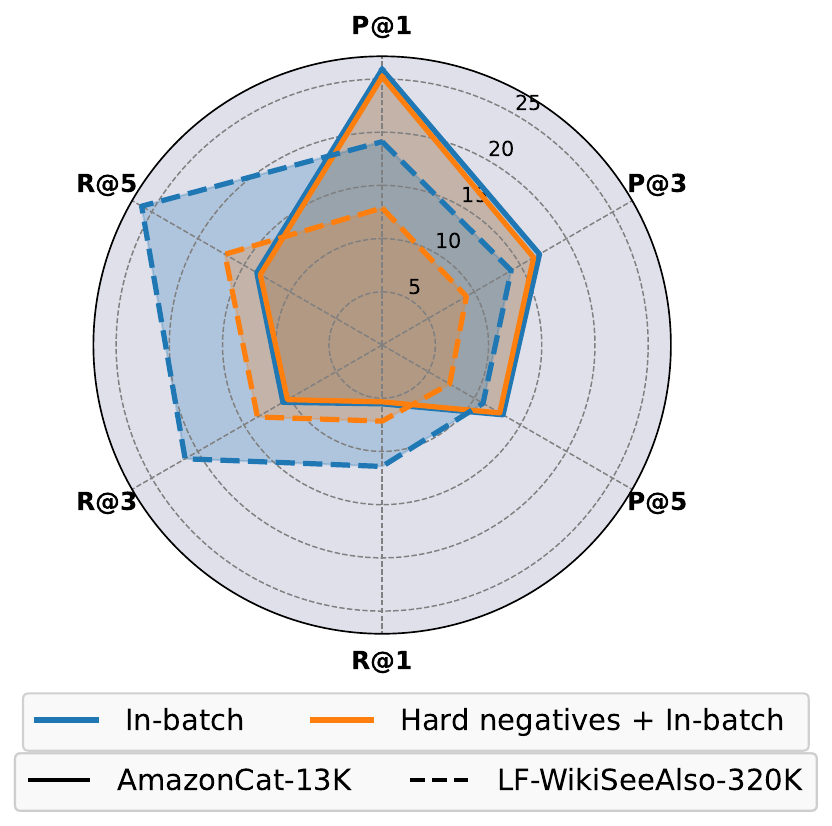}
  \caption{Comparative impact of negative sampling strategies on precision and recall performance.}
  \label{fig:nsampling}
\end{figure}

\noindent \textbf{Evaluating Negative Sampling and the Impact of LLM-Derived Hard Negatives:} Our bi-encoder training employs in-batch negatives. We extended this approach by incorporating hard negatives, identified by the LLM model and tagged as "no". For each document, we constructed a negative set comprising these hard negatives and the pseudo-positive labels of other documents within the same batch. Figure \ref{fig:nsampling} illustrates the comparative performance of these strategies. Notably, our results indicate that the inclusion of hard negatives can potentially impede bi-encoder training, likely due to the risk of introducing false negatives.

\section{Related Work}
\noindent \textbf{Supervised Extreme Multi-label Text Classification :} Supervised XMC methods leveraging non-label features include one-vs-rest approaches \cite{yen2016pd, babbar2017dismec, babbar2019data, schultheis2022speeding}, which are based on TF-IDF representations, as well as tree-based methods \cite{NEURIPS2019_9e6a921f, yu2022pecos, chang2020taming, jiang2021lightxml, liu2021label, gupta2022elias, zhang2021fast, kharbanda2022cascadexml, khandagale2020bonsai} that train distinct classifiers for different levels of the tree.
State-of-the-art non-label feature methods \cite{kharbanda2022cascadexml, zhang2021fast, jiang2021lightxml} are based on a transformer encoder and multi-layered tree classifiers. 
In contrast, state-of-the-art label feature methods \cite{saini2021galaxc, dahiya2021siamesexml, mittal2021eclare, dahiya2023ngame, mittal2021decaf, gupta2023efficacy} focus on embedding both label text and document text to achieve high accuracy.
All of these supervised approaches rely on well-annotated datasets and require comprehensive coverage of most of the labels in the training dataset. 

\noindent \textbf{Zero-shot Extreme Multi-label Text Classification:} The zero-shot XMC is capable of handling unseen labels which are not in the training dataset. ZestXML \cite{gupta2021generalized} applied generated TF-IDF features and  a trained linear model to enable retrieval of unseen labels.
However, this method still relies on well-annotated training datasets to learn the linear model and is not suitable for the cold start scenario. Another extreme setting in zero-shot XMC is Extreme Zero-shot Extreme Multi-label Text Classification (EZ-XMC) \cite{xiong-etal-2022-extreme}. EZ-XMC is specifically designed for the zero-shot scenario, particularly tailored for the cold start scenario without the need for a well-annotated training dataset. The key distinction between zero-shot XMC and EZ-XMC lies in whether annotated labels are employed in the training process. Unlike zero-shot XMC, EZ-XMC does not utilize any annotated labels. We adopt the EZ-XMC setting in this paper. 
MACLR \cite{xiong-etal-2022-extreme} proposes a multi-stage self-supervised approach for EZ-XMC by using pseudo pairs of (title, document).
On the other hand, RTS \cite{zhang-etal-2022-structural-contrastive} introduces a randomized text segmentation method to construct pseudo positive labels with segments within one document.

\noindent \textbf{Dense Sentence Embedding:} In the domains of open domain question answering and information retrieval, 
ICT (Inverse Cloze Task) \cite{lee2019latent} constructs positive passages by extracting random sentences and their corresponding contexts from the documents. 
MSS\cite{pmlr-v119-guu20a} shows that the ICT encoder can be improved by predicting the masked salient
spans with a reader. 
Spider \cite{ram2022learning} adopts sentences that contain recurring spans as positive passage. Both HLP \cite{zhou2022hyperlink} and WLP\cite{chang2019pre} utilize hyperlinks within Wikipedia pages to construct positive passages.
ART \cite{sachan2023questions} tries to guide the training of bi-encoder via the question reconstruction score.
Additionally, there are works that focus on sentence similarity, including (i) SimCSE \cite{gao2021simcse} introduces a contrastive learning framework that employs dropout noise as augmented positives, and (ii) 
Sentence-BERT \cite{reimers2019sentence} introduces a supervised siamese transformer framework.

\noindent \textbf{Large Language Models XMC Applications:} LLM models such as GPT-3 \cite{brown2020language}, and GPT-4 \cite{OpenAI2023GPT4TR} have demonstrated their zero-shot effectiveness in various NLP downstream tasks. 
In XMC, \citet{xu2023dense} employed LLM to construct a thesaurus for labels in a few-shot setting. \citet{liu2024xmc} applied LLM for   incremental XMC setting. \citet{zhu2024icxml}, on the other hand, directly applied the LLM for inference in EZ-XMC setting. This approach predominantly focuses on recommendation datasets and relies on the costly GPT-3.5 and GPT-4 for inference. In contrast, our methods concentrate on tagging tasks and emphasize swift inference through a lightweight bi-encoder.

\section{Conclusion}
This paper introduces a novel approach to address the EZ-XMC tagging and categorization challenge. We leverage an LLM as a teacher to guide the training of the bi-encoder model. Unlike existing methods, our approach effectively handles the issue of low-quality training pairs. Additionally, our algorithm enables faster inference without the need for an LLM during prediction, providing a significant advantage over LLM-based methods and supporting lightweight deployment in EZ-XMC scenarios. Performance evaluations demonstrate that our method achieves state-of-the-art results across multiple datasets. Ablation experiments further highlight its potential for improved performance when using alternative teacher models. For future work, exploring more efficient ways to integrate the LLM model is interesting, such as transitioning from point-wise to list-wise prompts, could be an exciting direction. 
\section{Limitations}
While our method demonstrates superior performance with a smaller subset, there is potential for further improvements with a larger training set (Figure \ref{fig:ablation1}). However, our current LLM pseudo-labeling approach relies on point-wise feedback, which is time-consuming. For the comparison with ICXML, we employed publicly available open-source models instead of GPT-3.5, which is specified in the original ICXML implementation. Despite this, benchmarking on large-label datasets proved computationally prohibitive. Instead, we used a subset of the test set and repeated the experiments multiple times to ensure statistical significance.
\section*{Acknowledgements}
We thank reviewers for their valuable comments and suggestions, we also sincerely thank Ansh Arora for his assistance in evaluating certain baselines of Zero-shot XMC models. We acknowledge the support of Research Council of Finland (Academy of Finland) via grants 347707 and 348215. We also thank the Aalto Science-IT project, and CSC IT Center for Science, Finland for the computational resources provided. 
\bibliography{custom}

\begin{thebibliography}{68}
\providecommand{\natexlab}[1]{#1}

\bibitem[{Babbar and Sch{\"o}lkopf(2017)}]{babbar2017dismec}
Rohit Babbar and Bernhard Sch{\"o}lkopf. 2017.
\newblock Dismec: Distributed sparse machines for extreme multi-label classification.
\newblock In \emph{Proceedings of the tenth ACM international conference on web search and data mining}, pages 721--729.

\bibitem[{Babbar and Sch{\"o}lkopf(2019)}]{babbar2019data}
Rohit Babbar and Bernhard Sch{\"o}lkopf. 2019.
\newblock Data scarcity, robustness and extreme multi-label classification.
\newblock \emph{Machine Learning}, 108(8-9):1329--1351.

\bibitem[{Bhatia et~al.(2016)Bhatia, Dahiya, Jain, Kar, Mittal, Prabhu, and Varma}]{Bhatia16}
K.~Bhatia, K.~Dahiya, H.~Jain, P.~Kar, A.~Mittal, Y.~Prabhu, and M.~Varma. 2016.
\newblock \href {http://manikvarma.org/downloads/XC/XMLRepository.html} {The extreme classification repository: Multi-label datasets and code}.

\bibitem[{Bonifacio et~al.(2022)Bonifacio, Abonizio, Fadaee, and Nogueira}]{bonifacio2022inpars}
Luiz Bonifacio, Hugo Abonizio, Marzieh Fadaee, and Rodrigo Nogueira. 2022.
\newblock Inpars: Data augmentation for information retrieval using large language models.
\newblock \emph{arXiv preprint arXiv:2202.05144}.

\bibitem[{Brown et~al.(2020)Brown, Mann, Ryder, Subbiah, Kaplan, Dhariwal, Neelakantan, Shyam, Sastry, Askell et~al.}]{brown2020language}
Tom Brown, Benjamin Mann, Nick Ryder, Melanie Subbiah, Jared~D Kaplan, Prafulla Dhariwal, Arvind Neelakantan, Pranav Shyam, Girish Sastry, Amanda Askell, et~al. 2020.
\newblock Language models are few-shot learners.
\newblock \emph{Advances in neural information processing systems}, 33:1877--1901.

\bibitem[{Chang et~al.(2019)Chang, Felix, Chang, Yang, and Kumar}]{chang2019pre}
Wei-Cheng Chang, X~Yu Felix, Yin-Wen Chang, Yiming Yang, and Sanjiv Kumar. 2019.
\newblock Pre-training tasks for embedding-based large-scale retrieval.
\newblock In \emph{International Conference on Learning Representations}.

\bibitem[{Chang et~al.(2021{\natexlab{a}})Chang, Jiang, Yu, Teo, Zhang, Zhong, Kolluri, Hu, Shandilya, Ievgrafov et~al.}]{chang2021extreme}
Wei-Cheng Chang, Daniel Jiang, Hsiang-Fu Yu, Choon~Hui Teo, Jiong Zhang, Kai Zhong, Kedarnath Kolluri, Qie Hu, Nikhil Shandilya, Vyacheslav Ievgrafov, et~al. 2021{\natexlab{a}}.
\newblock Extreme multi-label learning for semantic matching in product search.
\newblock In \emph{Proceedings of the 27th ACM SIGKDD Conference on Knowledge Discovery \& Data Mining}, pages 2643--2651.

\bibitem[{Chang et~al.(2021{\natexlab{b}})Chang, Jiang, Yu, Teo, Zhang, Zhong, Kolluri, Hu, Shandilya, Ievgrafov, Singh, and Dhillon}]{ChangJYTZZKHSIS21}
Wei{-}Cheng Chang, Daniel~L. Jiang, Hsiang{-}Fu Yu, Choon{-}Hui Teo, Jiong Zhang, Kai Zhong, Kedarnath Kolluri, Qie Hu, Nikhil Shandilya, Vyacheslav Ievgrafov, Japinder Singh, and Inderjit~S. Dhillon. 2021{\natexlab{b}}.
\newblock \href {https://doi.org/10.1145/3447548.3467092} {Extreme multi-label learning for semantic matching in product search}.
\newblock In \emph{{KDD} '21: The 27th {ACM} {SIGKDD} Conference on Knowledge Discovery and Data Mining, Virtual Event, Singapore, August 14-18, 2021}, pages 2643--2651. {ACM}.

\bibitem[{Chang et~al.(2020)Chang, Yu, Zhong, Yang, and Dhillon}]{chang2020taming}
Wei-Cheng Chang, Hsiang-Fu Yu, Kai Zhong, Yiming Yang, and Inderjit~S Dhillon. 2020.
\newblock Taming pretrained transformers for extreme multi-label text classification.
\newblock In \emph{Proceedings of the 26th ACM SIGKDD international conference on knowledge discovery \& data mining}, pages 3163--3171.

\bibitem[{Chiang et~al.(2023)Chiang, Li, Lin, Sheng, Wu, Zhang, Zheng, Zhuang, Zhuang, Gonzalez, Stoica, and Xing}]{vicuna2023}
Wei-Lin Chiang, Zhuohan Li, Zi~Lin, Ying Sheng, Zhanghao Wu, Hao Zhang, Lianmin Zheng, Siyuan Zhuang, Yonghao Zhuang, Joseph~E. Gonzalez, Ion Stoica, and Eric~P. Xing. 2023.
\newblock \href {https://lmsys.org/blog/2023-03-30-vicuna/} {Vicuna: An open-source chatbot impressing gpt-4 with 90\%* chatgpt quality}.

\bibitem[{Dahiya et~al.(2021)Dahiya, Agarwal, Saini, Gururaj, Jiao, Singh, Agarwal, Kar, and Varma}]{dahiya2021siamesexml}
Kunal Dahiya, Ananye Agarwal, Deepak Saini, K~Gururaj, Jian Jiao, Amit Singh, Sumeet Agarwal, Purushottam Kar, and Manik Varma. 2021.
\newblock Siamesexml: Siamese networks meet extreme classifiers with 100m labels.
\newblock In \emph{International Conference on Machine Learning}, pages 2330--2340. PMLR.

\bibitem[{Dahiya et~al.(2023)Dahiya, Gupta, Saini, Soni, Wang, Dave, Jiao, K, Dey, Singh et~al.}]{dahiya2023ngame}
Kunal Dahiya, Nilesh Gupta, Deepak Saini, Akshay Soni, Yajun Wang, Kushal Dave, Jian Jiao, Gururaj K, Prasenjit Dey, Amit Singh, et~al. 2023.
\newblock Ngame: Negative mining-aware mini-batching for extreme classification.
\newblock In \emph{Proceedings of the Sixteenth ACM International Conference on Web Search and Data Mining}, pages 258--266.

\bibitem[{Dai et~al.(2023)Dai, Shao, Zhao, Yu, Si, Xu, Sun, Zhang, and Xu}]{dai2023uncovering}
Sunhao Dai, Ninglu Shao, Haiyuan Zhao, Weijie Yu, Zihua Si, Chen Xu, Zhongxiang Sun, Xiao Zhang, and Jun Xu. 2023.
\newblock Uncovering chatgpt's capabilities in recommender systems.
\newblock \emph{arXiv preprint arXiv:2305.02182}.

\bibitem[{Dubey et~al.(2024)Dubey, Jauhri, Pandey, Kadian, Al-Dahle, Letman, Mathur, Schelten, Yang, Fan et~al.}]{dubey2024llama}
Abhimanyu Dubey, Abhinav Jauhri, Abhinav Pandey, Abhishek Kadian, Ahmad Al-Dahle, Aiesha Letman, Akhil Mathur, Alan Schelten, Amy Yang, Angela Fan, et~al. 2024.
\newblock The llama 3 herd of models.
\newblock \emph{arXiv preprint arXiv:2407.21783}.

\bibitem[{Gao et~al.(2021)Gao, Yao, and Chen}]{gao2021simcse}
Tianyu Gao, Xingcheng Yao, and Danqi Chen. 2021.
\newblock Simcse: Simple contrastive learning of sentence embeddings.
\newblock In \emph{Proceedings of the 2021 Conference on Empirical Methods in Natural Language Processing}, pages 6894--6910.

\bibitem[{Gupta et~al.(2021)Gupta, Bohra, Prabhu, Purohit, and Varma}]{gupta2021generalized}
Nilesh Gupta, Sakina Bohra, Yashoteja Prabhu, Saurabh Purohit, and Manik Varma. 2021.
\newblock Generalized zero-shot extreme multi-label learning.
\newblock In \emph{Proceedings of the 27th ACM SIGKDD Conference on Knowledge Discovery \& Data Mining}, pages 527--535.

\bibitem[{Gupta et~al.(2022)Gupta, Chen, Yu, Hsieh, and Dhillon}]{gupta2022elias}
Nilesh Gupta, Patrick Chen, Hsiang-Fu Yu, Cho-Jui Hsieh, and Inderjit Dhillon. 2022.
\newblock Elias: End-to-end learning to index and search in large output spaces.
\newblock \emph{Advances in Neural Information Processing Systems}, 35:19798--19809.

\bibitem[{Gupta et~al.(2023)Gupta, Devvrit, Rawat, Bhojanapalli, Jain, and Dhillon}]{gupta2023efficacy}
Nilesh Gupta, Fnu Devvrit, Ankit~Singh Rawat, Srinadh Bhojanapalli, Prateek Jain, and Inderjit~S Dhillon. 2023.
\newblock Efficacy of dual-encoders for extreme multi-label classification.
\newblock In \emph{The Twelfth International Conference on Learning Representations}.

\bibitem[{Guu et~al.(2020)Guu, Lee, Tung, Pasupat, and Chang}]{pmlr-v119-guu20a}
Kelvin Guu, Kenton Lee, Zora Tung, Panupong Pasupat, and Mingwei Chang. 2020.
\newblock \href {https://proceedings.mlr.press/v119/guu20a.html} {Retrieval augmented language model pre-training}.
\newblock In \emph{Proceedings of the 37th International Conference on Machine Learning}, volume 119 of \emph{Proceedings of Machine Learning Research}, pages 3929--3938. PMLR.

\bibitem[{Hou et~al.(2023)Hou, Zhang, Lin, Lu, Xie, McAuley, and Zhao}]{hou2023large}
Yupeng Hou, Junjie Zhang, Zihan Lin, Hongyu Lu, Ruobing Xie, Julian McAuley, and Wayne~Xin Zhao. 2023.
\newblock Large language models are zero-shot rankers for recommender systems.
\newblock \emph{arXiv preprint arXiv:2305.08845}.

\bibitem[{Jain et~al.(2016)Jain, Prabhu, and Varma}]{jain2016extreme}
Himanshu Jain, Yashoteja Prabhu, and Manik Varma. 2016.
\newblock Extreme multi-label loss functions for recommendation, tagging, ranking \& other missing label applications.
\newblock In \emph{Proceedings of the 22nd ACM SIGKDD international conference on knowledge discovery and data mining}, pages 935--944.

\bibitem[{Jiang et~al.(2021)Jiang, Wang, Sun, Yang, Zhao, and Zhuang}]{jiang2021lightxml}
Ting Jiang, Deqing Wang, Leilei Sun, Huayi Yang, Zhengyang Zhao, and Fuzhen Zhuang. 2021.
\newblock Lightxml: Transformer with dynamic negative sampling for high-performance extreme multi-label text classification.
\newblock In \emph{Proceedings of the AAAI Conference on Artificial Intelligence}, volume~35, pages 7987--7994.

\bibitem[{Johnson et~al.(2019)Johnson, Douze, and J{\'e}gou}]{johnson2019billion}
Jeff Johnson, Matthijs Douze, and Herv{\'e} J{\'e}gou. 2019.
\newblock Billion-scale similarity search with {GPUs}.
\newblock \emph{IEEE Transactions on Big Data}, 7(3):535--547.

\bibitem[{Khandagale et~al.(2020)Khandagale, Xiao, and Babbar}]{khandagale2020bonsai}
Sujay Khandagale, Han Xiao, and Rohit Babbar. 2020.
\newblock Bonsai: diverse and shallow trees for extreme multi-label classification.
\newblock \emph{Machine Learning}, 109:2099--2119.

\bibitem[{Kharbanda et~al.(2022)Kharbanda, Banerjee, Schultheis, and Babbar}]{kharbanda2022cascadexml}
Siddhant Kharbanda, Atmadeep Banerjee, Erik Schultheis, and Rohit Babbar. 2022.
\newblock Cascadexml: Rethinking transformers for end-to-end multi-resolution training in extreme multi-label classification.
\newblock \emph{Advances in Neural Information Processing Systems}, 35:2074--2087.

\bibitem[{Lee et~al.(2019)Lee, Chang, and Toutanova}]{lee2019latent}
Kenton Lee, Ming-Wei Chang, and Kristina Toutanova. 2019.
\newblock Latent retrieval for weakly supervised open domain question answering.
\newblock In \emph{Proceedings of the 57th Annual Meeting of the Association for Computational Linguistics}, pages 6086--6096.

\bibitem[{Liu et~al.(2017)Liu, Chang, Wu, and Yang}]{liu2017deep}
Jingzhou Liu, Wei-Cheng Chang, Yuexin Wu, and Yiming Yang. 2017.
\newblock Deep learning for extreme multi-label text classification.
\newblock In \emph{Proceedings of the 40th international ACM SIGIR conference on research and development in information retrieval}, pages 115--124.

\bibitem[{Liu et~al.(2021)Liu, Chang, Yu, Hsieh, and Dhillon}]{liu2021label}
Xuanqing Liu, Wei-Cheng Chang, Hsiang-Fu Yu, Cho-Jui Hsieh, and Inderjit Dhillon. 2021.
\newblock Label disentanglement in partition-based extreme multilabel classification.
\newblock \emph{Advances in Neural Information Processing Systems}, 34:15359--15369.

\bibitem[{Liu et~al.(2024)Liu, Zhong, Lu, Lin, He, Zhou, Zhu, Wang, Liu, Han et~al.}]{liu2024xmc}
Yanjiang Liu, Tianyun Zhong, Yaojie Lu, Hongyu Lin, Ben He, Shuheng Zhou, Huijia Zhu, Weiqiang Wang, Zhongyi Liu, Xianpei Han, et~al. 2024.
\newblock Xmc-agent: Dynamic navigation over scalable hierarchical index for incremental extreme multi-label classification.
\newblock In \emph{Findings of the Association for Computational Linguistics ACL 2024}, pages 5659--5672.

\bibitem[{Loshchilov and Hutter(2018)}]{loshchilov2018decoupled}
Ilya Loshchilov and Frank Hutter. 2018.
\newblock Decoupled weight decay regularization.
\newblock In \emph{International Conference on Learning Representations}.

\bibitem[{Ma et~al.(2023)Ma, Zhang, Pradeep, and Lin}]{ma2023zero}
Xueguang Ma, Xinyu Zhang, Ronak Pradeep, and Jimmy Lin. 2023.
\newblock Zero-shot listwise document reranking with a large language model.
\newblock \emph{arXiv preprint arXiv:2305.02156}.

\bibitem[{Malkov and Yashunin(2018)}]{malkov2018efficient}
Yu~A Malkov and Dmitry~A Yashunin. 2018.
\newblock Efficient and robust approximate nearest neighbor search using hierarchical navigable small world graphs.
\newblock \emph{IEEE transactions on pattern analysis and machine intelligence}, 42(4):824--836.

\bibitem[{Manmatha et~al.(2017)Manmatha, Wu, Smola, and Kr{\"a}henb{\"u}hl}]{Manmatha2017SamplingMI}
R.~Manmatha, Chaoxia Wu, Alex Smola, and Philipp Kr{\"a}henb{\"u}hl. 2017.
\newblock \href {https://api.semanticscholar.org/CorpusID:24718057} {Sampling matters in deep embedding learning}.
\newblock \emph{2017 IEEE International Conference on Computer Vision (ICCV)}, pages 2859--2867.

\bibitem[{Mittal et~al.(2021{\natexlab{a}})Mittal, Dahiya, Agrawal, Saini, Agarwal, Kar, and Varma}]{mittal2021decaf}
Anshul Mittal, Kunal Dahiya, Sheshansh Agrawal, Deepak Saini, Sumeet Agarwal, Purushottam Kar, and Manik Varma. 2021{\natexlab{a}}.
\newblock Decaf: Deep extreme classification with label features.
\newblock In \emph{Proceedings of the 14th ACM International Conference on Web Search and Data Mining}, pages 49--57.

\bibitem[{Mittal et~al.(2021{\natexlab{b}})Mittal, Sachdeva, Agrawal, Agarwal, Kar, and Varma}]{mittal2021eclare}
Anshul Mittal, Noveen Sachdeva, Sheshansh Agrawal, Sumeet Agarwal, Purushottam Kar, and Manik Varma. 2021{\natexlab{b}}.
\newblock Eclare: Extreme classification with label graph correlations.
\newblock In \emph{Proceedings of the Web Conference 2021}, pages 3721--3732.

\bibitem[{OpenAI(2023)}]{OpenAI2023GPT4TR}
OpenAI. 2023.
\newblock \href {https://api.semanticscholar.org/CorpusID:257532815} {Gpt-4 technical report}.
\newblock \emph{ArXiv}, abs/2303.08774.

\bibitem[{Pennington et~al.(2014)Pennington, Socher, and Manning}]{pennington-etal-2014-glove}
Jeffrey Pennington, Richard Socher, and Christopher Manning. 2014.
\newblock \href {https://doi.org/10.3115/v1/D14-1162} {{G}lo{V}e: Global vectors for word representation}.
\newblock In \emph{Proceedings of the 2014 Conference on Empirical Methods in Natural Language Processing ({EMNLP})}, pages 1532--1543, Doha, Qatar. Association for Computational Linguistics.

\bibitem[{Qaraei et~al.(2021)Qaraei, Schultheis, Gupta, and Babbar}]{qaraei2021convex}
Mohammadreza Qaraei, Erik Schultheis, Priyanshu Gupta, and Rohit Babbar. 2021.
\newblock Convex surrogates for unbiased loss functions in extreme classification with missing labels.
\newblock In \emph{Proceedings of the Web Conference 2021}, pages 3711--3720.

\bibitem[{Qin et~al.(2023)Qin, Jagerman, Hui, Zhuang, Wu, Shen, Liu, Liu, Metzler, Wang et~al.}]{qin2023large}
Zhen Qin, Rolf Jagerman, Kai Hui, Honglei Zhuang, Junru Wu, Jiaming Shen, Tianqi Liu, Jialu Liu, Donald Metzler, Xuanhui Wang, et~al. 2023.
\newblock Large language models are effective text rankers with pairwise ranking prompting.
\newblock \emph{arXiv preprint arXiv:2306.17563}.

\bibitem[{Ram et~al.(2022)Ram, Shachaf, Levy, Berant, and Globerson}]{ram2022learning}
Ori Ram, Gal Shachaf, Omer Levy, Jonathan Berant, and Amir Globerson. 2022.
\newblock Learning to retrieve passages without supervision.
\newblock In \emph{Proceedings of the 2022 Conference of the North American Chapter of the Association for Computational Linguistics: Human Language Technologies}, pages 2687--2700.

\bibitem[{Reddi et~al.(2019)Reddi, Kale, Yu, Holtmann-Rice, Chen, and Kumar}]{pmlr-v89-reddi19a}
Sashank~J. Reddi, Satyen Kale, Felix Yu, Daniel Holtmann-Rice, Jiecao Chen, and Sanjiv Kumar. 2019.
\newblock Stochastic negative mining for learning with large output spaces.
\newblock In \emph{Proceedings of the Twenty-Second International Conference on Artificial Intelligence and Statistics}, volume~89 of \emph{Proceedings of Machine Learning Research}, pages 1940--1949. PMLR.

\bibitem[{Reimers and Gurevych(2019)}]{reimers2019sentence}
Nils Reimers and Iryna Gurevych. 2019.
\newblock Sentence-bert: Sentence embeddings using siamese bert-networks.
\newblock In \emph{Proceedings of the 2019 Conference on Empirical Methods in Natural Language Processing and the 9th International Joint Conference on Natural Language Processing (EMNLP-IJCNLP)}, pages 3982--3992.

\bibitem[{Reimers and Gurevych(2021)}]{reimers2021curse}
Nils Reimers and Iryna Gurevych. 2021.
\newblock The curse of dense low-dimensional information retrieval for large index sizes.
\newblock In \emph{Proceedings of the 59th Annual Meeting of the Association for Computational Linguistics and the 11th International Joint Conference on Natural Language Processing (Volume 2: Short Papers)}, pages 605--611.

\bibitem[{Saad-Falcon et~al.(2023)Saad-Falcon, Khattab, Santhanam, Florian, Franz, Roukos, Sil, Sultan, and Potts}]{saad2023udapdr}
Jon Saad-Falcon, Omar Khattab, Keshav Santhanam, Radu Florian, Martin Franz, Salim Roukos, Avirup Sil, Md~Arafat Sultan, and Christopher Potts. 2023.
\newblock Udapdr: Unsupervised domain adaptation via llm prompting and distillation of rerankers.
\newblock \emph{arXiv preprint arXiv:2303.00807}.

\bibitem[{Sachan et~al.(2022)Sachan, Lewis, Joshi, Aghajanyan, Yih, Pineau, and Zettlemoyer}]{sachan-etal-2022-improving}
Devendra Sachan, Mike Lewis, Mandar Joshi, Armen Aghajanyan, Wen-tau Yih, Joelle Pineau, and Luke Zettlemoyer. 2022.
\newblock \href {https://aclanthology.org/2022.emnlp-main.249} {Improving passage retrieval with zero-shot question generation}.
\newblock In \emph{Proceedings of the 2022 Conference on Empirical Methods in Natural Language Processing}, pages 3781--3797, Abu Dhabi, United Arab Emirates. Association for Computational Linguistics.

\bibitem[{Sachan et~al.(2023)Sachan, Lewis, Yogatama, Zettlemoyer, Pineau, and Zaheer}]{sachan2023questions}
Devendra~Singh Sachan, Mike Lewis, Dani Yogatama, Luke Zettlemoyer, Joelle Pineau, and Manzil Zaheer. 2023.
\newblock Questions are all you need to train a dense passage retriever.
\newblock \emph{Transactions of the Association for Computational Linguistics}, 11:600--616.

\bibitem[{Saini et~al.(2021)Saini, Jain, Dave, Jiao, Singh, Zhang, and Varma}]{saini2021galaxc}
Deepak Saini, Arnav~Kumar Jain, Kushal Dave, Jian Jiao, Amit Singh, Ruofei Zhang, and Manik Varma. 2021.
\newblock Galaxc: Graph neural networks with labelwise attention for extreme classification.
\newblock In \emph{Proceedings of the Web Conference 2021}, pages 3733--3744.

\bibitem[{Sanh et~al.(2019)Sanh, Debut, Chaumond, and Wolf}]{sanh2019distilbert}
Victor Sanh, Lysandre Debut, Julien Chaumond, and Thomas Wolf. 2019.
\newblock Distilbert, a distilled version of bert: smaller, faster, cheaper and lighter.
\newblock \emph{arXiv preprint arXiv:1910.01108}.

\bibitem[{Schroff et~al.(2015{\natexlab{a}})Schroff, Kalenichenko, and Philbin}]{Schroff_2015_CVPR}
Florian Schroff, Dmitry Kalenichenko, and James Philbin. 2015{\natexlab{a}}.
\newblock Facenet: A unified embedding for face recognition and clustering.
\newblock In \emph{Proceedings of the IEEE Conference on Computer Vision and Pattern Recognition (CVPR)}.

\bibitem[{Schroff et~al.(2015{\natexlab{b}})Schroff, Kalenichenko, and Philbin}]{schroff2015facenet}
Florian Schroff, Dmitry Kalenichenko, and James Philbin. 2015{\natexlab{b}}.
\newblock Facenet: A unified embedding for face recognition and clustering.
\newblock In \emph{Proceedings of the IEEE conference on computer vision and pattern recognition}, pages 815--823.

\bibitem[{Schultheis and Babbar(2021)}]{schultheis2021unbiased}
Erik Schultheis and Rohit Babbar. 2021.
\newblock Unbiased loss functions for multilabel classification with missing labels.
\newblock \emph{arXiv preprint arXiv:2109.11282}.

\bibitem[{Schultheis and Babbar(2022)}]{schultheis2022speeding}
Erik Schultheis and Rohit Babbar. 2022.
\newblock Speeding-up one-versus-all training for extreme classification via mean-separating initialization.
\newblock \emph{Machine Learning}, 111(11):3953--3976.

\bibitem[{Schultheis et~al.(2022)Schultheis, Wydmuch, Babbar, and Dembczynski}]{schultheis2022missing}
Erik Schultheis, Marek Wydmuch, Rohit Babbar, and Krzysztof Dembczynski. 2022.
\newblock On missing labels, long-tails and propensities in extreme multi-label classification.
\newblock In \emph{Proceedings of the 28th ACM SIGKDD Conference on Knowledge Discovery and Data Mining}, pages 1547--1557.

\bibitem[{Schultheis et~al.(2024)Schultheis, Wydmuch, Kotlowski, Babbar, and Dembczynski}]{schultheis2024generalized}
Erik Schultheis, Marek Wydmuch, Wojciech Kotlowski, Rohit Babbar, and Krzysztof Dembczynski. 2024.
\newblock Generalized test utilities for long-tail performance in extreme multi-label classification.
\newblock \emph{Advances in Neural Information Processing Systems}, 36.

\bibitem[{Song et~al.(2020)Song, Tan, Qin, Lu, and Liu}]{song2020mpnet}
Kaitao Song, Xu~Tan, Tao Qin, Jianfeng Lu, and Tie-Yan Liu. 2020.
\newblock Mpnet: Masked and permuted pre-training for language understanding.
\newblock \emph{Advances in Neural Information Processing Systems}, 33:16857--16867.

\bibitem[{Sun et~al.(2023)Sun, Yan, Ma, Ren, Yin, and Ren}]{sun2023chatgpt}
Weiwei Sun, Lingyong Yan, Xinyu Ma, Pengjie Ren, Dawei Yin, and Zhaochun Ren. 2023.
\newblock Is chatgpt good at search? investigating large language models as re-ranking agent.
\newblock \emph{arXiv preprint arXiv:2304.09542}.

\bibitem[{Touvron et~al.(2023)Touvron, Martin, Stone, Albert, Almahairi, Babaei, Bashlykov, Batra, Bhargava, Bhosale et~al.}]{touvron2023llama}
Hugo Touvron, Louis Martin, Kevin Stone, Peter Albert, Amjad Almahairi, Yasmine Babaei, Nikolay Bashlykov, Soumya Batra, Prajjwal Bhargava, Shruti Bhosale, et~al. 2023.
\newblock Llama 2: Open foundation and fine-tuned chat models.
\newblock \emph{arXiv preprint arXiv:2307.09288}.

\bibitem[{Wydmuch et~al.(2021)Wydmuch, Jasinska-Kobus, Babbar, and Dembczynski}]{wydmuch2021propensity}
Marek Wydmuch, Kalina Jasinska-Kobus, Rohit Babbar, and Krzysztof Dembczynski. 2021.
\newblock Propensity-scored probabilistic label trees.
\newblock In \emph{Proceedings of the 44th International ACM SIGIR Conference on Research and Development in Information Retrieval}, pages 2252--2256.

\bibitem[{Xiong et~al.(2022)Xiong, Chang, Hsieh, Yu, and Dhillon}]{xiong-etal-2022-extreme}
Yuanhao Xiong, Wei-Cheng Chang, Cho-Jui Hsieh, Hsiang-Fu Yu, and Inderjit Dhillon. 2022.
\newblock \href {https://doi.org/10.18653/v1/2022.naacl-main.399} {Extreme {Z}ero-{S}hot learning for extreme text classification}.
\newblock In \emph{Proceedings of the 2022 Conference of the North American Chapter of the Association for Computational Linguistics: Human Language Technologies}, pages 5455--5468, Seattle, United States. Association for Computational Linguistics.

\bibitem[{Xu et~al.(2023{\natexlab{a}})Xu, Sun, Zheng, Geng, Zhao, Feng, Tao, and Jiang}]{xu2023wizardlm}
Can Xu, Qingfeng Sun, Kai Zheng, Xiubo Geng, Pu~Zhao, Jiazhan Feng, Chongyang Tao, and Daxin Jiang. 2023{\natexlab{a}}.
\newblock \href {https://arxiv.org/abs/2304.12244} {Wizardlm: Empowering large language models to follow complex instructions}.
\newblock \emph{Preprint}, arXiv:2304.12244.

\bibitem[{Xu et~al.(2023{\natexlab{b}})Xu, Wang, Dong, and Chen}]{xu2023dense}
Nan Xu, Fei Wang, Mingtao Dong, and Muhao Chen. 2023{\natexlab{b}}.
\newblock Dense retrieval as indirect supervision for large-space decision making.
\newblock \emph{arXiv preprint arXiv:2310.18619}.

\bibitem[{Yen et~al.(2016)Yen, Huang, Ravikumar, Zhong, and Dhillon}]{yen2016pd}
Ian En-Hsu Yen, Xiangru Huang, Pradeep Ravikumar, Kai Zhong, and Inderjit Dhillon. 2016.
\newblock Pd-sparse: A primal and dual sparse approach to extreme multiclass and multilabel classification.
\newblock In \emph{International conference on machine learning}, pages 3069--3077. PMLR.

\bibitem[{You et~al.(2019)You, Zhang, Wang, Dai, Mamitsuka, and Zhu}]{NEURIPS2019_9e6a921f}
Ronghui You, Zihan Zhang, Ziye Wang, Suyang Dai, Hiroshi Mamitsuka, and Shanfeng Zhu. 2019.
\newblock \href {https://proceedings.neurips.cc/paper_files/paper/2019/file/9e6a921fbc428b5638b3986e365d4f21-Paper.pdf} {Attentionxml: Label tree-based attention-aware deep model for high-performance extreme multi-label text classification}.
\newblock In \emph{Advances in Neural Information Processing Systems}, volume~32. Curran Associates, Inc.

\bibitem[{Yu et~al.(2022)Yu, Zhong, Zhang, Chang, and Dhillon}]{yu2022pecos}
Hsiang-Fu Yu, Kai Zhong, Jiong Zhang, Wei-Cheng Chang, and Inderjit~S Dhillon. 2022.
\newblock Pecos: Prediction for enormous and correlated output spaces.
\newblock \emph{the Journal of machine Learning research}, 23(1):4233--4264.

\bibitem[{Zhang et~al.(2021)Zhang, Chang, Yu, and Dhillon}]{zhang2021fast}
Jiong Zhang, Wei-Cheng Chang, Hsiang-Fu Yu, and Inderjit Dhillon. 2021.
\newblock Fast multi-resolution transformer fine-tuning for extreme multi-label text classification.
\newblock \emph{Advances in Neural Information Processing Systems}, 34:7267--7280.

\bibitem[{Zhang et~al.(2022)Zhang, Xu, Medini, and Shrivastava}]{zhang-etal-2022-structural-contrastive}
Tianyi Zhang, Zhaozhuo Xu, Tharun Medini, and Anshumali Shrivastava. 2022.
\newblock \href {https://aclanthology.org/2022.findings-emnlp.362} {Structural contrastive representation learning for zero-shot multi-label text classification}.
\newblock In \emph{Findings of the Association for Computational Linguistics: EMNLP 2022}, pages 4937--4947, Abu Dhabi, United Arab Emirates. Association for Computational Linguistics.

\bibitem[{Zhou et~al.(2022)Zhou, Li, Shang, Luo, Zhan, Hu, Zhang, Jiang, Cao, Yu et~al.}]{zhou2022hyperlink}
Jiawei Zhou, Xiaoguang Li, Lifeng Shang, Lan Luo, Ke~Zhan, Enrui Hu, Xinyu Zhang, Hao Jiang, Zhao Cao, Fan Yu, et~al. 2022.
\newblock Hyperlink-induced pre-training for passage retrieval in open-domain question answering.
\newblock In \emph{Proceedings of the 60th Annual Meeting of the Association for Computational Linguistics (Volume 1: Long Papers)}, pages 7135--7146.

\bibitem[{Zhu and Zamani(2024)}]{zhu2024icxml}
Yaxin Zhu and Hamed Zamani. 2024.
\newblock Icxml: An in-context learning framework for zero-shot extreme multi-label classification.
\newblock In \emph{Findings of the Association for Computational Linguistics: NAACL 2024}, pages 2086--2098.

\end{thebibliography}

\appendix

\section{Appendix}
\label{sec:appendix}

\subsection{Implementation Details}
\label{app:implementation}
\textbf{Bi-Encoder:} In our bi-encoder framework, we adopt a siamese network architecture for sentence encoding. The core of this network is DistilBERT \cite{sanh2019distilbert}, comprising six transformer layers. For the generation of sentence embeddings, we apply mean pooling, yielding embeddings of 768 dimensions.
The bi-encoder is initialized using the msmarco-distilbert-base-v4\footnote{https://huggingface.co/sentence-transformers/msmarco-distilbert-base-v4}, and  ANNs is built via the HNSW package\footnote{https://github.com/kunaldahiya/pyxclib}. For optimization, we employ the AdamW optimizer \cite{loshchilov2018decoupled} with a learning rate of 0.0002, setting the batch size to 128. All experiments for training the bi-encoder are conducted on a single A100 GPU. Following the supervised method in \cite{dahiya2023ngame}, we have used triplet loss \cite{schroff2015facenet,liu2017deep} with margin $\gamma$ is set to 0.3.  For model selection, a development set of 800 documents is randomly selected from the training dataset, with pseudo labels derived from the top-k labels as determined by the LLM model. 

\noindent \textbf{LLM:} For our Large Language Model (LLM) component, we employ the WizardLM-13B-V1.0 model \cite{xu2023wizardlm}, an open-source LLM notable for achieving 89.1\% of GPT-4's \cite{OpenAI2023GPT4TR} performance with approximately 13 billion parameters. In addition, for the purposes of this study, we incorporate Llama2 \cite{touvron2023llama} and vicuna-13b-v1.3 \cite{vicuna2023} models in our ablation experiments to serve as comparative benchmarks.  All LLM computations are performed on 2 $\times$ A100 GPUs, with input instances truncated to 430 tokens. For the comparison with ICXML \cite{zhu2024icxml}, we adopt Llama3 \cite{dubey2024llama} and vicuna-33b-v1.3 \cite{vicuna2023} for inference.

\noindent \textbf{Random Training Subsets:} To minimize bias from random subsets for AmazonCat-13K, LF-WikiSeeAlso-320K, and LF-Wikipedia-500K, we conducted three separate random samplings and used the average performance of the three models on the test set as our final result in Table \ref{tab:LMTX_performance}.

\subsection{Evaluation Metrics} 
\label{app:evaluation_metrics} We employ the commonly used evaluation metrics \cite{pmlr-v89-reddi19a,ChangJYTZZKHSIS21, zhang-etal-2022-structural-contrastive} for the EZ-XMC setting : $Precision@k (P@m)$ and $Recall@m (R@m)$. 
\begin{equation}
\footnotesize
P@m = \frac{1}{m}\sum_{i \in rank_{m}(\hat{y})}y_{i} \text{,}\quad R@m = \frac{1}{\sum_{l}y_{l}}\sum_{i \in rank_{m}(\hat{y})}y_{i}
\end{equation}
where $\hat{y} \in \mathbb{R}^{L}$ represents a vector containing the predicted labels' score for each instance, while $y \in \{0, 1\}^L$ corresponds to a vector representing the ground truth for each document. The term $rank_m(\hat{y})$ refers to a list of the predicted top-$m$ label indices. The definition of the two metrics applies to a single instance; for multiple instances, the performance is the average across all instances.

\begin{table}[h]
\small
\centering
\resizebox{\linewidth}{!}{%
\begin{tabular}{c|c|cl}
\hline
\textbf{Dataset} & \textbf{Models} & \textbf{Training} & \textbf{GPUs} \\ \hline
\multirow{3}{*}{\begin{tabular}[c]{@{}c@{}}AmazonCat- \\ 13K\end{tabular}} & MACLR & 28.86 & 4 A100 \\
 & RTS & 35.60 & 4 A100 \\
 & LMTX 30k & \textbf{22.79} & 2 A100 \\ \hline
\multirow{3}{*}{\begin{tabular}[c]{@{}c@{}}LF- \\ WikiSeeAlso-\\ 320K\end{tabular}} & MACLR & 28.88 & 4 A100 \\
 & RTS & 26.66 & 4 A100 \\
 & LMTX 30k & \textbf{26.03} & 2 A100 \\ \hline
\end{tabular}%
}
\caption{The training time (in hours) comparison with non-LLM methods.}
\label{tab:training time}
\end{table}

\subsection{Training time} In Table \ref{tab:training time}, we present the training time for our model when  trained with a subset of the training set. The table shows that LMTX's time efficiency is competitive or even superior compared to other models, especially in the context of larger datasets. These results underscore the effectiveness of LMTX,  even with the incorporation of the LLM model.

\begin{figure}[ht]
  \centering
  \includegraphics[width=\linewidth]{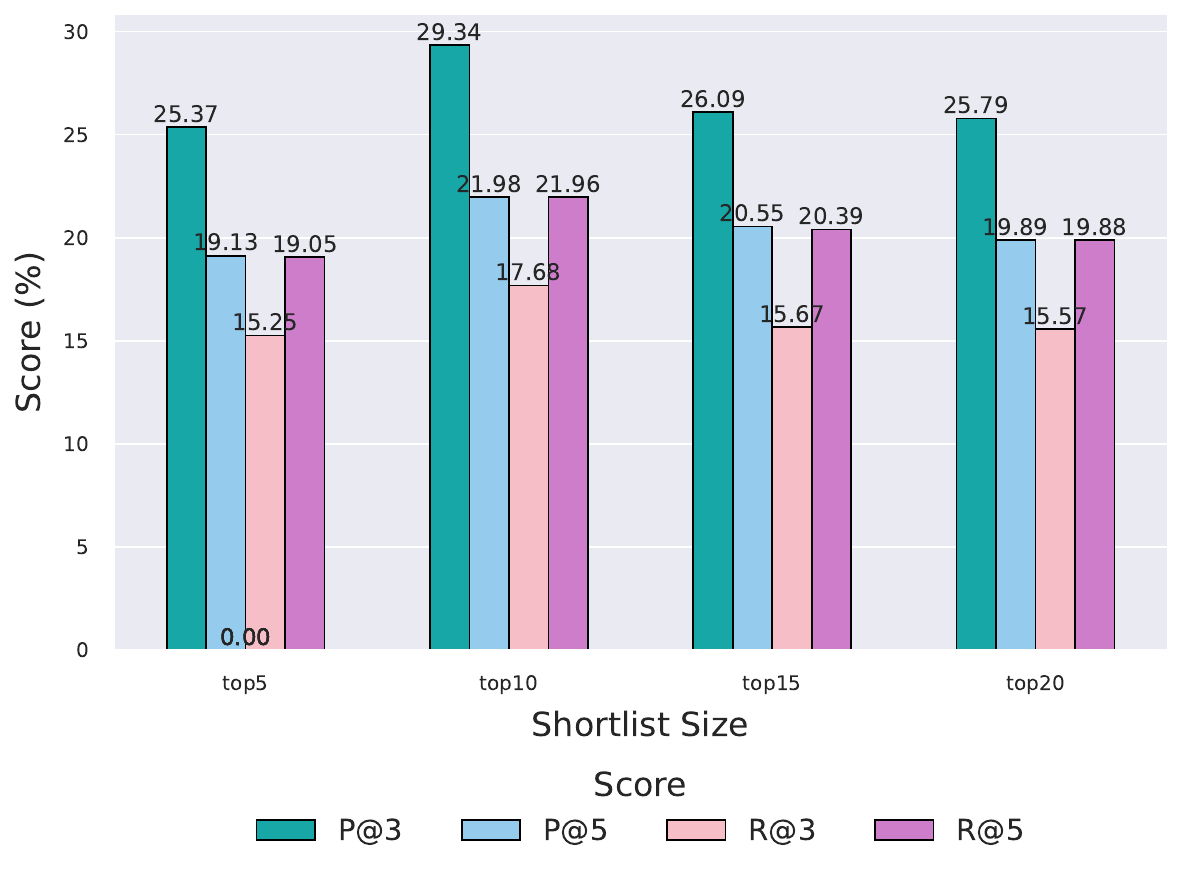}
  \caption{Impact of shortlist size on performance metrics for EURLex-4k dataset. }
  \label{fig:topk}
\end{figure}

\subsection{Sensitivity to the Shortlist Size} The size of the shortlist directly impacts both the quality of pseudo labels generated by the LLM and the computational efficiency of the screening process. We empirically evaluated the effect of varying shortlist sizes on precision and recall for the EURLex-4K dataset, as illustrated in \autoref{fig:topk}. Our results demonstrate that while a shortlist size of 5 negatively impacts performance, increasing the size beyond 10 does not yield significant improvements. Notably, we observed optimal performance across multiple metrics at a shortlist size of 10, indicating that our approach achieves superior results with a relatively compact shortlist, thereby enhancing training efficiency.

\subsection{Evaluating Pseudo-Label Quality and the Role of Curriculum Learning:} 
\begin{figure}[ht]
  \centering
  \includegraphics[width=\linewidth]{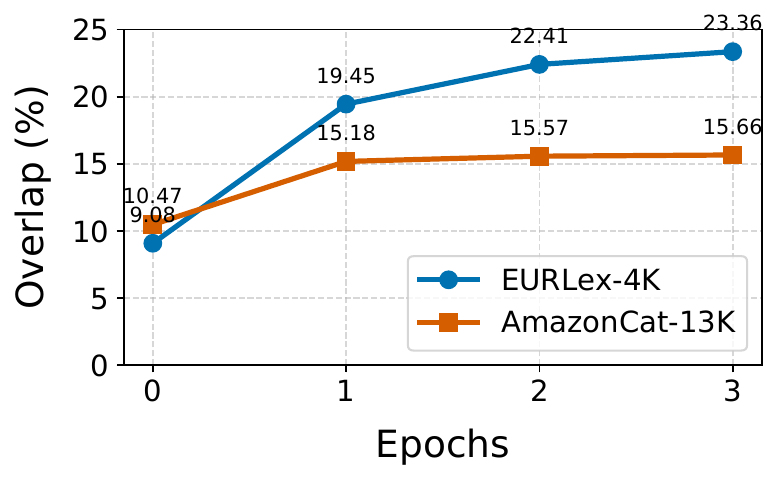}
  \caption{Overlap ratio between LLM-generated pseudo-labels and ground truth labels across training epochs for EURLex-4K and AmazonCat-13K datasets.}
  \label{fig:overlap}
\end{figure}

To assess the LLM's capability in selecting relevant labels and the quality of the selected  pseudo-labels, we measured the overlap between the pseudo-labels and the supervised ground truth. The overlap ratio is calculated as follows

$$
quality = \frac{len(pseudo\_labels \cap true\_labels) }{len(true\_labels)} 
$$
As illustrated in Figure \ref{fig:overlap}, the overlap ratio progressively increases across training epochs for both the EURLex-4K and AmazonCat-13K datasets. This trend demonstrates the effectiveness of our curriculum learning framework, as the LLM refines its label selection over time, resulting in higher-quality pseudo-labels. The increasing overlap highlights that the curriculum learning strategy not only improves pseudo-label alignment with ground truth but also enhances the performance of the bi-encoder. 

\subsection{Prompts for LLM}\label{prompts_llm}
\begin{itemize}
\item  \textbf{EURLex-4k}  and \textbf{Wiki10-31K}: “document = \{doc\}. Is the tag \{label\_text\} relevant to the document? answer yes or no”

\item \textbf{AmazonCat-13K}: “document = \{doc\}. The document is amazon product description, Is the tag \{label\_text\} relevant to the document? answer yes or no”

\item \textbf{LF-WikiSeeAlso-320K}: "document = \{doc\}. The document is the wikipedia page. Does another wikipedia page name "\{label\_text\}" has the relation to the document? answer yes or no"

\item \textbf{LF-Wikipedia-500K}:"document = \{doc\}, the document is the wikipedia page. Is the tag "\{label\_text\}" relevant to the document? answer yes or no".
\end{itemize}

\end{document}